\documentclass[letterpaper, 10 pt, conference]{ieeeconf}  

\IEEEoverridecommandlockouts                              
\makeatletter
\let\NAT@parse\undefined
\makeatother

\usepackage[dvipsnames]{xcolor}

\newcommand*\linkcolours{ForestGreen}

\usepackage{times}
\usepackage{graphicx}
\usepackage{amssymb}
\usepackage{gensymb}
\usepackage{amsmath}
\usepackage{breakurl}

\usepackage{url,hyperref}
\hypersetup{
colorlinks,
linkcolor=\linkcolours,
citecolor=\linkcolours,
filecolor=\linkcolours,
urlcolor=\linkcolours}

\usepackage{algorithm}
\usepackage{algorithmic}

\usepackage[labelfont={bf},font=small]{caption}
\usepackage[none]{hyphenat}

\usepackage{mathtools, cuted}

\usepackage[noadjust, nobreak]{cite}

\usepackage{tabularx}
\usepackage{amsmath}

\usepackage{float}

\usepackage{pifont}

\newcommand*\GitHubLoc{https://github.com/Jeffrey-Ede/ALRC}

\newcolumntype{Y}{>{\centering\arraybackslash}X}

\usepackage[]{placeins}

\newcommand\extraspace{3pt}

\usepackage{placeins}

\usepackage{tikz}

\usepackage[framemethod=tikz]{mdframed}

\usepackage{afterpage}

\usepackage{stfloats}

\usepackage{atbegshi}
\newcommand{\handlethispage}{}
\newcommand{\discardpagesfromhere}{\let\handlethispage\AtBeginShipoutDiscard}
\newcommand{\keeppagesfromhere}{\let\handlethispage\relax}
\AtBeginShipout{\handlethispage}

\usepackage{comment}

\title{\LARGE \bf
Adaptive Learning Rate Clipping Stabilizes Learning
}

\author{Jeffrey M. Ede$^{1}$ and Richard Beanland$^{2}$
\thanks{$^{1}$Jeffrey is a PhD student in Materials and Analytical Sciences, 
University of Warwick, CV4 7AL
Email: j.m.ede@warwick.ac.uk}%
\thanks{$^{2}$Richard is a Reader in the Deparment of Physics, 
University of Warwick, CV4 7AL
Email: r.beanland@warwick.ac.uk}%
}

\begin{document}

\maketitle
\thispagestyle{empty}
\pagestyle{empty}

\begin{abstract}

Artificial neural network training with stochastic gradient descent can be destabilized by ``bad batches'' with high losses. This is often problematic for training with small batch sizes, high order loss functions or unstably high learning rates. To stabilize learning, we have developed adaptive learning rate clipping (ALRC) to limit backpropagated losses to a number of standard deviations above their running means. ALRC is designed to complement existing learning algorithms: Our algorithm is computationally inexpensive, can be applied to any loss function or batch size, is robust to hyperparameter choices and does not affect backpropagated gradient distributions. Experiments with CIFAR-10 supersampling show that ALCR decreases errors for unstable mean quartic error training while stable mean squared error training is unaffected. We also show that ALRC decreases unstable mean squared errors for partial scanning transmission electron micrograph completion. Our source code is publicly available at \url{\GitHubLoc}.

\end{abstract}

\section{INTRODUCTION}

\begin{figure}[tbp]
\centering
\includegraphics[width=0.97\columnwidth]{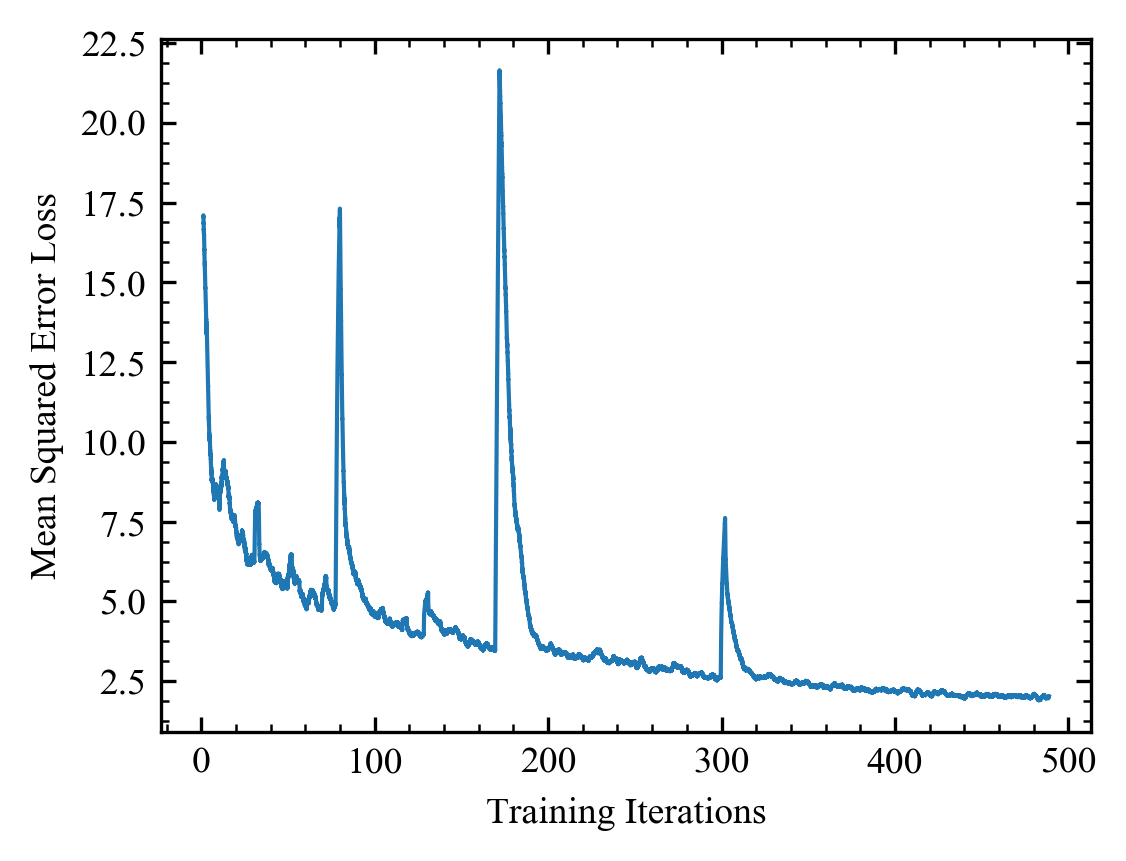}
\caption{ Learning curve with high loss spikes that excessively perturb a trainable parameter distribution. Losses decrease after loss spikes as parameters are updated back to an intelligent distribution. The learning curve is 2500 iteration boxcar averaged. }
\label{loss_spikes}
\end{figure}

This paper addresses loss spikes, one of the most common reasons for low performance in artificial neural networks trained with stochastic gradient descent\cite{ruder2016overview} (SGD). Gradients backpropagated from high losses can excessively perturb trainable parameter distributions and destabilize learning. An example of loss spikes destabilizing learning is shown in fig.~\ref{loss_spikes}. Loss spikes are common for small batch sizes, high order loss functions and unstably high learning rates.

During neural network training with vanilla SGD, a trainable parameter, $\theta_t$, from step $t$ is updated to $\theta_{t+1}$ in step $t+1$. The size of the update is given by the product of a learning rate, $\eta$, and the backpropagated gradient of a loss function with respect to the trainable parameter
\begin{equation}
\theta_{t+1} \leftarrow \theta_t - \eta \dfrac{\partial L}{\partial \theta}.
\end{equation}
Without modification, trainable parameter perturbations are proportional to the scale of the loss function. This means that a loss spike will cause a large perturbation to the learned parameter distribution. Learning will then be destabilized while subsequent iterations update trainable parameters back to an intelligent distribution. 

Trainable parameter perturbations are often limited by clipping gradients to a multiple of their global norm\cite{bengio2012difficulty}. For large batch sizes, this can limit perturbations by loss spikes as their gradients will be larger than other gradients in the batch. However, global norm clipping alters the distribution of gradients backpropagated from high losses and is unable to identify and clip high losses if the batch size is too small. Clipping gradients of individual layers by their norms has the same limitations. 

Gradient clipping to a user-provided threshold can also be applied globally or to individual layers. This can limit loss spike perturbations for any batch size. However, the clipping threshold is an extra hyperparameter to determine and may need to be changed throughout training. Further, it does not preserve distributions of gradients for high losses. 

More commonly, destabilizing perturbations are reduced by selecting a low order loss function and stable learning rate. Low order loss functions; such as absolute and squared distances, are effective because they are less prone to destabilizingly high errors than higher-order loss functions. Indeed, loss function modifications used to stabilize learning often lower loss function order. For instance, Huberization \cite{huber1964robust} reduces perturbations by losses, $L$, larger than $h$ by applying the mapping $L \rightarrow \min(L,(hL)^{1/2})$.

\section{Algorithm}

Adaptive learning rate clipping (ALRC, algorithm~\ref{alrc_algorithm}) is designed to addresses the limitations of gradient clipping. Namely, to be computationally inexpensive, effective for any batch size, robust to hyperparameter choices and to preserve backpropagated gradient distributions. Like gradient clipping, it also has to be applicable to arbitrary loss funtions and neural network architectures.  

\begin{algorithm}[h]
\caption{Adaptive learning rate clipping (ALRC) of loss spikes. Sensible parameters are $\beta_1 = \beta_2 = 0.999$, $n=3$ and $\mu_1^2 < \mu_2$.}
\begin{algorithmic}
\STATE Initialize running means, $\mu_1$ and $\mu_2$, with decay rates, $\beta_1$ and $\beta_2$.
\STATE Choose number, $n$, of standard deviations to clip to.
\WHILE{Training is not finished}
  \STATE Infer forward-propagation loss, $L$.
  \STATE $\sigma \leftarrow (\mu_2 - \mu_1^2)^{1/2}$ 
  \STATE $L_\text{max} \leftarrow \mu_1 + n \sigma$
  \IF{$L > L_\text{max}$}
  \STATE $L_\text{dyn} \leftarrow \text{stop\_gradient}(L_\text{max}/L)L$
  \ELSE\STATE $L_\text{dyn} \leftarrow L$
  \ENDIF
  \STATE Optimize network by back-propagating $L_\text{dyn}$.
  \STATE $\mu_1 \leftarrow \beta_1 \mu_1 + (1-\beta_1) L$
  \STATE $\mu_2 \leftarrow \beta_2 \mu_2 + (1-\beta_2) L^2$
\ENDWHILE
\end{algorithmic}
\label{alrc_algorithm}
\end{algorithm}

Rather than allowing loss spikes to destabilize learning, ALRC applies the mapping $\eta L \rightarrow \text{stop\_gradient}(L_\text{max}/L) \eta L$ if $L > L_\text{max}$. The function $\text{stop\_gradient}$ leaves its operand unchanged in the forward pass and blocks gradients in the backwards pass. ALRC adapts the learning rate to limit the effective loss being backpropagated to $L_\text{max}$. The value of $L_\text{max}$ is non-trivial for ALRC to complement existing learning algorithms. In addition to training stability and robustness to hyperparameter choices, $L_\text{max}$ needs to adapt to losses and learning rates as they vary. 

In our implementation, $L_\text{max}$ is a number of standard deviations of the loss above its mean and requires five hyperparameters. There are two decay rates, $\beta_1$ and $\beta_2$, for exponential moving averages used to estimate the mean and standard deviation of the loss and a number, $n$, of standard deviations. Similar to batch normalization\cite{ioffe2015batch}, any decay rate close to 1 is effective e.g. $\beta_1 = \beta_2 = 0.999$. Performance does vary slightly with $n$; however, we find that any $n \approx 3$ is effective. Initial values for the running means, $\mu_1$ and $\mu_2$, where $\mu_1^2 < \mu_2$ also have to be provided. However, any sensible initial estimates larger than their true values are fine as $\mu_1$ and $\mu_2$ will decay to their correct values.

ALRC can be extended to any loss function or batch size. For batch sizes above 1, we apply ALRC to individual losses, while $\mu_1$ and $\mu_2$ are updated with mean losses. ARLC can also be applied to loss summands; such as per pixel errors between generated and reference images, while $\mu_1$ and $\mu_2$ are updated with the mean errors.

\section{Experiments: CIFAR-10 Supersampling}

To invistagate the ability of ALRC to stabilize learning and its robustness to hyperperameter choices, we performed a series of toy experiments with networks trained to upsample CIFAR-10\cite{krizhevsky2014cifar, krizhevsky2009learning} images to 32$\times$32$\times$3 after downsampling to 16$\times$16$\times$3.

\vspace{\extraspace}
\noindent\textbf{Data pipeline:} In order, images were randomly flipped left or right, had their brightness altered, had their contrast altered, were linearly transformed to have zero mean and unit variance and bilinearly downsampled to 16$\times$16$\times$3.

\begin{figure}[tbh!]
\centering
\includegraphics[width=0.57\columnwidth]{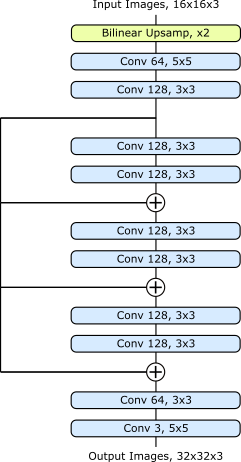}
\caption{ Convolutional image 2$\times$ supersampling network with three skip-2 residual blocks. }
\label{alrc_architecture}
\end{figure}

\vspace{\extraspace}
\noindent\textbf{Architecture:} Images were upsampled and passed through the convolutional network in fig.~\ref{alrc_architecture}. Each convolutional layer is followed by ReLU\cite{nair2010rectified} activation, except the last. 

\vspace{\extraspace}
\noindent\textbf{Initialization:} All weights were Xavier\cite{glorot2010understanding} initialized. Biases were zero initialized.

\vspace{\extraspace}
\noindent\textbf{Learning policy:} ADAM optimization was used with the hyperparameters recommended in \cite{kingma2014adam} and a base learning rate of 1/1280 for 100000 iterations. The learning rate was constant in batch size 1, 4, 16 experiments and decreased to 1/12800 after 54687 iterations in batch size 64 experiments. Networks were trained to minimize mean squared or quartic errors between restored and ground truth images. ALRC was applied to limit the magnitudes of losses to either 2, 3, 4 or $\infty$ standard deviations above their running means. For batch sizes above 1, ALRC was applied to each loss individually.

\begin{figure*}[tbh!]
\vspace{0.7cm}
{\centering
\includegraphics[width=\textwidth]{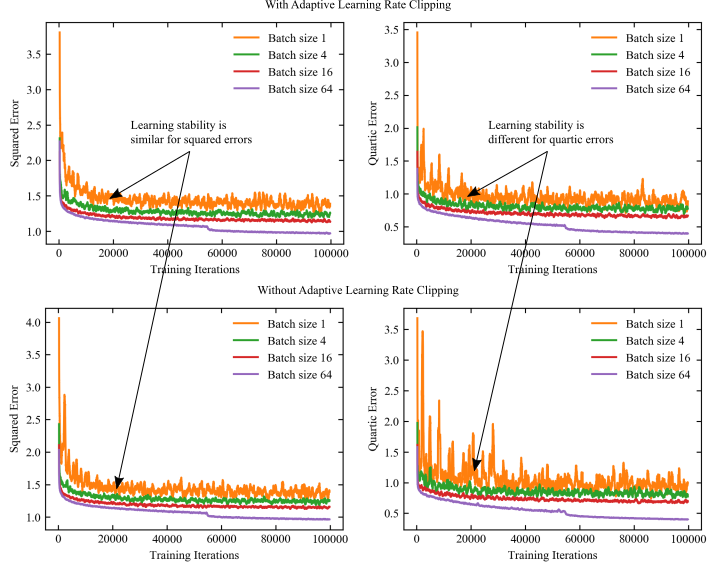}
\caption{ Unclipped learning curves for 2$\times$ CIFAR-10 upsampling with batch sizes 1, 4, 16 and 64 with and without adaptive learning rate clipping of losses to 3 standard deviations above their running means. Training is more stable for squared errors than quartic errors. Learning curves are 500 iteration boxcar averaged. }
\label{fig:alrc}}
\footnotesize
\vspace{\baselineskip}
Squared Errors\\
\begin{tabular*}{\textwidth}{@{\extracolsep{\fill}}c|cccccccc}
\hline
\multicolumn{1}{c|}{}       & \multicolumn{2}{c}{Batch Size 1} & \multicolumn{2}{c}{Batch Size 4} & \multicolumn{2}{c}{Batch Size 16} & \multicolumn{2}{c}{Batch Size 64} \\
Threshold & Mean         & \multicolumn{1}{c}{Std Dev}      & Mean       & Std Dev      & Mean       & \multicolumn{1}{c}{Std Dev}       & Mean      & Std Dev      \\ \hline
2 & 5.55 & 0.048 & 4.96 & 0.016 & 4.58 & 0.010 & - & - \\
3 & 5.52 & 0.054 & 4.96 & 0.029 & 4.58 & 0.004 & 3.90 & 0.013 \\
4 & 5.56 & 0.048 & 4.97 & 0.017 & 4.58 & 0.007 & 3.89 & 0.016 \\
$\infty$ & 5.55 & 0.041 & 4.98 & 0.017 & 4.59 & 0.006 & 3.89 & 0.014 \\
\hline
\end{tabular*}
\vspace{\baselineskip}

Quartic Errors\\
\begin{tabular*}{\textwidth}{@{\extracolsep{\fill}}c|cccccccc}
\hline
\multicolumn{1}{c|}{}       & \multicolumn{2}{c}{Batch Size 1} & \multicolumn{2}{c}{Batch Size 4} & \multicolumn{2}{c}{Batch Size 16} & \multicolumn{2}{c}{Batch Size 64} \\
Threshold & Mean         & \multicolumn{1}{c}{Std Dev}      & Mean       & Std Dev      & Mean       & \multicolumn{1}{c}{Std Dev}       & Mean      & Std Dev      \\ \hline
2 & 3.54 & 0.084 & 3.02 & 0.023 & 2.60 & 0.012 & 1.65 & 0.011 \\
3 & 3.59 & 0.055 & 3.08 & 0.024 & 2.61 & 0.014 & 1.58 & 0.016 \\
4 & 3.61 & 0.054 & 3.13 & 0.023 & 2.64 & 0.016 & 1.57 & 0.016 \\
$\infty$ & 3.88 & 0.108 & 3.32 & 0.037 & 2.74 & 0.020 & 1.61 & 0.008 \\
\hline
\end{tabular*}
\captionof{table}{ Adaptive learning rate clipping (ALRC) for losses 2, 3, 4 and $\infty$ running standard deviations above their running means for batch sizes 1, 4, 16 and 64. ARLC was not applied for clipping at $\infty$. Each squared and quartic error mean and standard deviation is for the means of the final 5000 training errors of 10 experiments. ALRC lowers errors for unstable quartic error training at low batch sizes and otherwise has little effect. Means and standard deviations are multiplied by 100. }
\label{table:alrc}
\end{figure*}

\vspace{\extraspace}
\noindent\textbf{Results:} Example learning curves for mean squared and quartic error training are shown in fig.~\ref{fig:alrc}. Training is more stable and converges to lower losses for larger batch sizes. Training is less stable for quartic errors than squared errors, allowing ALRC to be examined for loss functions with different stability.

Training was repeated 10 times for each combination of ALRC threshold and batch size. Means and standard deviations of the means of the last 5000 training losses for each experiment are tabulated in table~\ref{table:alrc}. ALRC has no effect on mean squared error (MSE) training, even for batch size 1. However, it decreases errors for batch sizes 1, 4 and 16 for mean quartic error training.

\FloatBarrier
\discardpagesfromhere
\clearpage
\keeppagesfromhere

\begin{figure*}[tbh]
\centering
\includegraphics[width=0.93\textwidth]{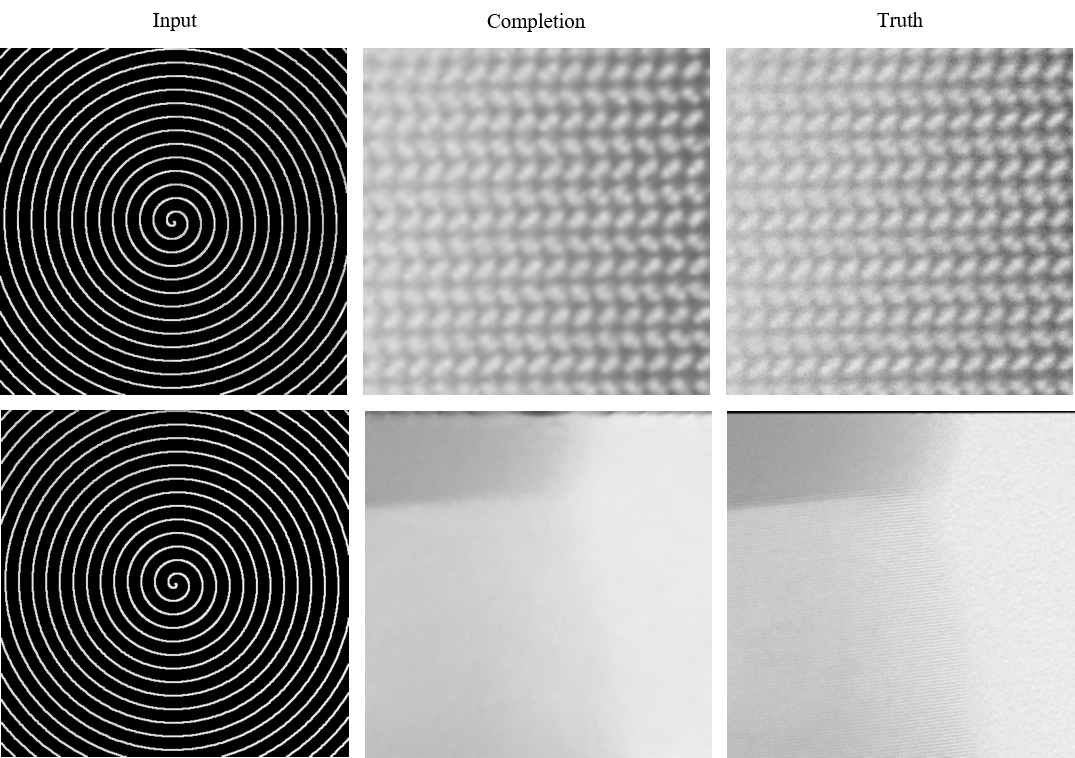}
\caption{ Neural network completions of 512$\times$512 scanning transmission electron microscopy images from 1/20 coverage blurred spiral scans. }
\label{alrc_example}
\end{figure*}

\section{Experiments: Partial-STEM}

To test ALRC in practice, we applied our algorithm to neural networks learning to complete 512$\times$512 scanning transmission electron microscopy (STEM) images from partial scans with 1/20 coverage. Example completions are shown in fig.~\ref{alrc_example}.

\begin{figure*}[tbp!]
\vspace{1.5cm}
\centering
\includegraphics[width=0.97\textwidth]{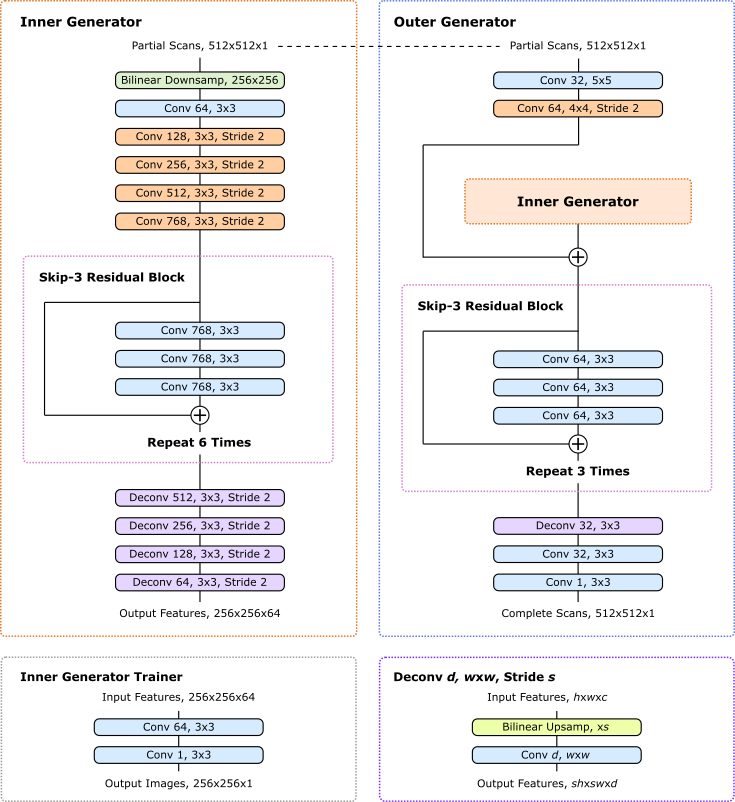}
\caption{ Two-stage generator that completes 512$\times$512 micrographs from partial scans. A dashed line indicates that the same image is input to the inner and outer generator. Large scale features developed by the inner generator are locally enhanced by the outer generator and turned into images. An auxiliary inner generator trainer restores images from inner generator features to provide direct feedback. }
\label{gen-2-step}
\vspace{1.5cm}
\end{figure*}

\vspace{\extraspace}
\noindent\textbf{Data pipeline:} In order, each image was subject to a random combination of flips and 90$\degree$ rotations to augment the dataset by a factor of 8. Next, each STEM images was blurred and a path described by a 1/20 coverage spiral was selected. Finally, artificial noise was added to scans to make them more difficult to complete.

\vspace{\extraspace}
\noindent\textbf{Architecture:} Our network can be divided into the three subnetworks shown in fig.~\ref{gen-2-step}: an inner generator, outer generator and an auxiliary inner generator trainer. The auxiliary trainer\cite{szegedy2014going, szegedy2015rethinking} is introduced to provide a more direct path for gradients to backpropagate to the inner generator. Each convolutional layer is followed by ReLU activation, except the last.

\vspace{\extraspace}
\noindent\textbf{Initialization:} Weights were initialized from a normal distribution with mean 0.00 and standard deviation 0.05. There are no biases.

\vspace{\extraspace}
\noindent\textbf{Weight normalization:} All generator weights are weight normalized\cite{salimans2016weight} and a weight normalization initialization pass was performed after weight initialization. Following \cite{salimans2016weight, hoffer2018norm}, running mean-only batch normalization was applied to the output channels of every convolutional layer except the last. Channel means were tracked by exponential moving averages with decay rates of 0.99. Similar to \cite{chen2017rethinking}, running mean-only batch normalization was frozen in the second half of training to improve stability.

\vspace{\extraspace}
\noindent\textbf{Loss functions:} The auxiliary inner generator trainer learns to generate half-size completions that minimize MSEs from half-size blurred ground truth STEM images. Meanwhile, the outer generator learns to produce full-size completions that minimize MSEs from blurred STEM images. All MSEs were multipled by 200. The inner generator cooperates with the auxiliary inner generator trainer and outer generator. 

To benchmark ALRC, we investigated training with MSEs, Huberized ($h=1$) MSEs, MSEs with ALRC and Huberized ($h=1$) MSEs with ALRC before Huberization. Training with both ALRC and Hubarization showcases the ability of ALRC to complement another loss function modification.

\vspace{\extraspace}
\noindent\textbf{Learning policy:} ADAM optimization\cite{kingma2014adam} was used with a constant generator learning rate of 0.0003 and a first moment of the momentum decay rate, $\beta_1=0.9$, for 250000 iterations. In the next 250000 iterations, the learning rate and $\beta_1$ were linearly decayed in eight steps to zero and 0.5, respectively. The learning rate for the auxiliary inner generator trainer was two times the generator learning rate; $\beta_1$ were the same. All training was performed with batch size 1 due to the large model size needed to complete 512$\times$512 scans.

\begin{figure}[htbp]
\centering
\includegraphics[width=0.97\columnwidth]{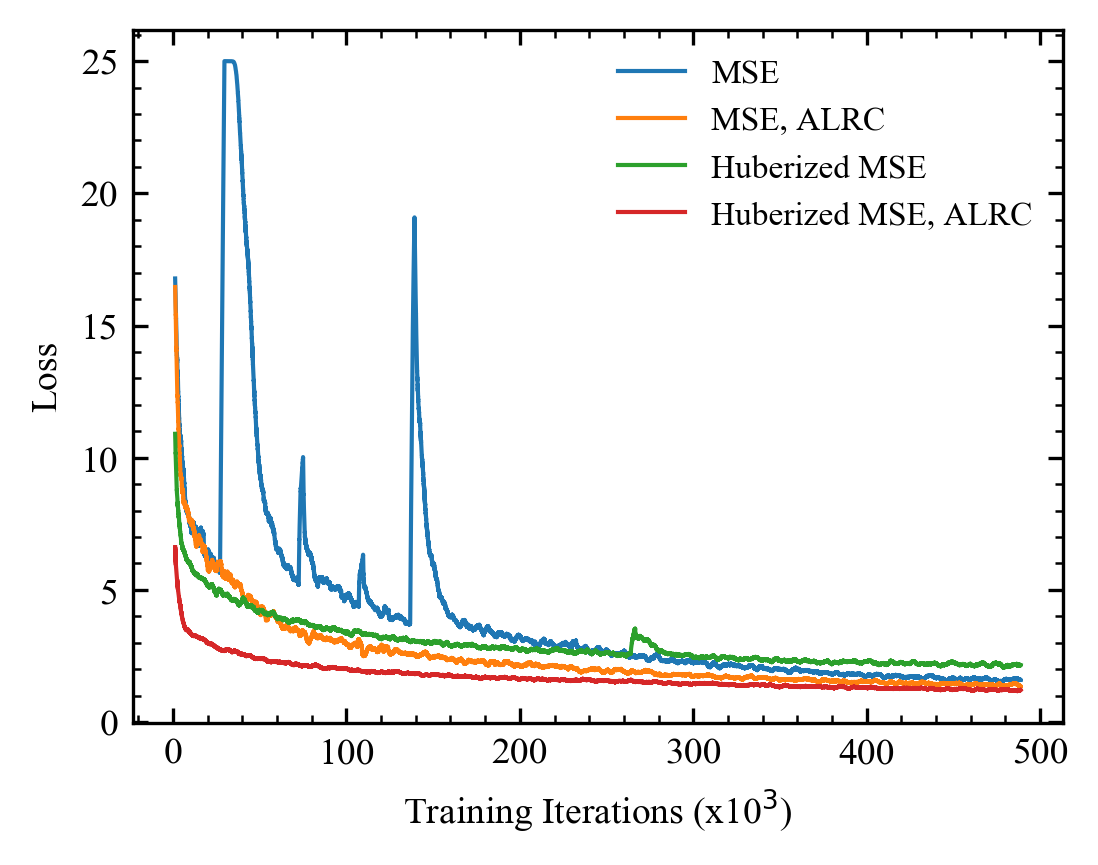}
\caption{ Outer generator losses show that ALRC and Huberization stabilize learning. ALRC lowers mean squared error (MSE) and Huberized MSE losses and accelerates convergence. Learning curves are 2500 iteration boxcar averaged. }
\label{stability}
\end{figure}

\vspace{\extraspace}
\noindent\textbf{Results:} Outer generator losses in fig.~\ref{stability} show that ALRC and Huberization stabilize learning. Further, ALRC accelerates MSE and Huberized MSE convergence to lower losses. To be clear, learning policy was optimized for MSE training so direct loss comparison is uncharitable to ALRC.

\section{Discussion}

Taken together, our CIFAR-10 supersampling results show that ALRC improves stability and lowers losses for learning that would be destabilized by loss spikes and otherwise has little effect. Loss spikes are often encountered when training with high learning rates, high order loss functions or small batch sizes. However, a moderate learning rate was used in MSE experiments so losses did not spike enough to destabilize learning. In contrast, mean quartic error training is unstable so ALRC stabilizes training and lowers losses. Similar results are confirmed for partial-STEM where ALRC stabilizes learning and lowers losses.

ALRC is designed to complement existing learning algorithms with new functionality. It is effective for any loss function or batch size and can be applied to any neural network trained with a variant of stochastic gradient descent. Our algorithm is also computationally inexpensive, requiring orders of magnitude fewer operations than other layers typically used in neural networks. As ALRC either stabilizes learning or has little effect, this means that it is suitable for routine application to arbitrary neural network training with SGD. In addition, we note that ALRC is a simple algorithm that has a clear effect on learning.

Nevertheless, ALRC can replace other learning algorithms in some situations. For instance, ALRC is a computationally inexpensive alternative to gradient clipping in high batch size training where gradient clipping is being used to limit perturbations by loss spikes. However, it is not a direct replacement as ALRC preserves the distribution of backpropagated gradients whereas gradient clipping reduces large gradients. Instead, ALRC is designed to complement gradient clipping by limiting perturbations by large losses while gradient clipping modifies gradient distributions.

The implementation of ALRC in algorithm~\ref{alrc_algorithm} is for positive losses. This avoids the need to introduce small constants to prevent divide-by-zero errors. Nevertheless, ALRC can support negative losses by using standard methods to prevent divide by zero errors. Alternatively, a constant can be added to losses to make them positive without affecting learning.

ALRC can also be extended to limit losses more than a number of standard deviations below their mean. This had no effect in our experiments. However, preemptively reducing loss spikes by clipping rewards between user-provided upper and lower bounds can improve reinforcement learning\cite{mnih2015human}. Subsequently, we suggest that clipping losses below their means did not improve learning because losses mainly spiked above their means; not below. Some partial-STEM losses did spike below; however, they were mainly for blank or otherwise trivial completions.

\section{Conclusions}

We have developed ALRC to stabilize the training of artificial neural networks by limiting backpropagated losses. Our experiments show that ALRC accelerates convergence and lowers losses for learning that would be destabilized by loss spikes and otherwise has little effect. Further, ALRC is computationally inexpensive, can be applied to any loss function or batch size, does not affect the distribution of backpropagated gradients and has a clear effect on learning. Overall, ALRC complements existing learning algorithms and can be routinely applied to arbitrary neural network training with SGD.

\section{Source Code}

Source code for CIFAR-10 supersampling experiments and a TensorFlow\cite{abadi2016tensorflow} implementation of ALRC is available at \url{\GitHubLoc}.



\bibliographystyle{ieeetr}
\bibliography{bibliography}

\section{Acknowledgements}

\noindent This research was funded by EPSRC grant EP/N035437/1.

\clearpage

\end{document}